\documentclass{sigplanconf}

\usepackage{graphicx}
\usepackage{amsmath}
\usepackage{hyperref}

\begin{document}

\setlength{\pdfpageheight}{\paperheight}
\setlength{\pdfpagewidth}{\paperwidth}

\conferenceinfo{CONF '24}{March 23, 2024, Islamabad, Pakistan}
\copyrightyear{2024}
\copyrightdata{978-1/24/03}
\doi{123.456}




\titlebanner{banner above paper title}        
\preprintfooter{short description of paper}   

\title{Enhancing Robustness in Biomedical NLI Models: A Probing Approach for Clinical Trials}
\subtitle{SemEval 2024 Task 2: Safe Biomedical Natural Language Inference for Clinical Trials}
\authorinfo{Ata Mustafa}
           {National University of \\ Computer \& Emerging Sciences,\\Islamabad}
           {i222943@nu.edu.pk}
\maketitle

\begin{abstract}
Large Language Models have revolutionized various fields and industries, such as Conversational AI, Content Generation, Information Retrieval, Business Intelligence, and Medical, to name a few. One major application in the field of medical is to analyze and investigate clinical trials for entailment tasks. However, It has been observed that Large Language Models are susceptible to shortcut learning, factual inconsistency, and performance degradation with little variation in context. Adversarial and robust testing is performed to ensure the integrity of models' output. But, ambiguity still persists. In order to ensure the integrity of the reasoning performed and investigate the model has correct syntactic and semantic understanding probing is used. Here, I used mnestic probing to investigate the SciFive model, trained on clinical trial. I investigated the model for feature learnt with respect to natural logic. To achieve the target, I trained task specific probes. Used these probes to investigate the final layers of trained model. Then, fine tuned the trained model using iterative null projection. The results shows that model accuracy improved. During experimentation, I observed that size of the probe has affect on the fine tuning process.  
\end{abstract}




\section{Problem statement}
Large Language Models (LLMs) are susceptible to shortcut learning, factual inconsistency, and performance degradation when exposed to word distribution shifts, data transformations, and adversarial examples. These limitations can lead to an overestimation of the real-world performance. These limitations are therefore of particular concern in the context of medical applications. The LLMs need to be investigated regarding their consistency in representation of semantic phenomena and their ability to perform faithful reasoning.
\section{Introduction}
Large Language Models (LLMs) have achieved state-of-the-art performance in many NLP fields like Natural Language Generation \& Understanding, Language Translation, Information retrieval. Expectations are models perform robust deductive reasoning, capture contextual information, deal with semantic variations, and find complex conceptual relations.  The developed models are found susceptible to short cut learning and factual inconsistency. There is adverse degradation in performance if there is a little data shift. The use of LLMs in vital fields like real-world clinical trials necessitates the creation of innovative evaluation methods rooted in more methodical behavioral and causal analyses \cite{xing2020tasty}. The given scenario rises to different questions. Which features of data are driving the model decisions? What will happen if the information is removed from the input data? What will be the affect of rephrasing input data? What will be the affect of logical and semantic variations.

\begin{figure}
    \centering
    \includegraphics[width=0.5\linewidth]{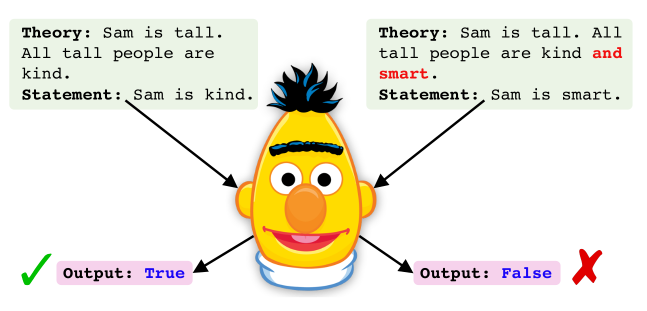}
    \caption{An example of miss classification}
    \label{fig:ex1}
\end{figure}

To eleborate the scenario a hypothectical model is shown in fig \ref{fig:ex1}. Logical variation is performed in the input of model and it predicted the false entailment.

Different NLI models for the clinical trials have been developed. One of the most prominent model is BioBERT (Bidirectional Encoder Representations from Transformers for Biomedical Text Mining), based on the BERT architecture. ClinicalBERT is trained on clinical notes. MedBERT is trained to capture the context and semantic of medical language. SciFive is another text-to-text transformer model in the domain. It is based on T5 model. The developed model have a heavy contirbution in the field of medical. But there is a big question about the integrity of the inference performed by the model. Hallucination problem is another major threat. Incorrect response from the models may lead to severe consequence. Consequences may range from degradation in treatment process to the death of someone. Need of the hour is to develop some mechanism to investigate the consistency of the models regarding their representation of semantic phenomena and their ability to perform faithful reasoning. 

In this work, I develop a framework to investigate the semantic correctness and robustness of state-of-the-art model using mnesting probing. I leverage rigorous settings of \textit{natural logic}. It is based on two complex features: \textit{monotonicity} and \textit{concept inclusion relation}. I fine tune the pretrained model, SciFive, on \textit{SemEval 2024 Task 2: Safe Biomedical Natural Language Inference for Clinical Trials} dataset. SciFive is a text-to-text transformer model based on T5. I investigate the fine tuned model to identify to which extent the model has captured the montonicity and concept inclusion relation feature in its internal representation. To do the job, I trained task specific probes. Using the probes, performed iterative null space projection on last layer of decoder part. I sort out the key performer feature for the entailment task using probing. This help in rectifying the superfluous features and fine tune the model. This results in a more robust and semantically correct decision maker model.

\section{Related work}

NLI model's semantic correctness task assesses the accuracy of the relationship between a premise and a hypothesis. Richardson et al. proposed the use of semantic fragments, systematically generated datasets targeting different semantic phenomena, to investigate the abilities of language understanding models beyond basic linguistic understanding \cite{richardson2020probing}. Sanyal et al. introduced a diagnostic benchmark that evaluates the robustness of language models in deductive reasoning by testing their ability to understand logical operators and handle logical perturbations. They analyzed RoBERTa, T5, and GPT3 for their robustness and found that the models struggle with logical reasoning \cite{sanyal2022robustlr}.

Faithful reasoning determines the NLI Model integrity. To investigate the faithfulness Xing et al. developed a new test set called Aspect Robustness Test Set (ARTS) and analyze the aspect robustness of nine ABSA models, finding a significant drop in accuracy \cite{xing2020tasty}. Yu et al. suggested a new training method, Bottom-Up Automatic Intervention (BAI), to help NLU models perform better in different, unfamiliar situations. BAI intervenes at multiple levels to address the issue. It's shown to be effective in experiments on three NLU tasks: language understanding, fact checking, and recognizing paraphrases \cite{yu2022interventional}. Rozanova et al. applied causal effect estimation strategies to measure the effect of context interventions and interventions on the inserted word-pair. They also compare model profiles in terms of robustness to irrelevant changes and sensitivity to desired changes \cite{rozanova2023estimating}.

According to Roznaova et al. simple data intervention techniques are not enough to guarantee that the model is trained on intended features \cite{rozanova2023interventional}. Elazer et al. introduced amnesic probing to perform the model intervention technique \cite{elazar2021amnesic}. The technique modeled the features to perform the downstream task. The technique is based on iterative nullspace projection(INLP) \cite{ravfogel2020null}. Efficiency of amnesic probing is limited when number of data dimension is high and number of classes is low. To deal with the issue, Roznaova et al. modified the technique and float the idea of mnestic probing based on natural logic. The idea is to keep only directions identified by the iteratively trained during INLP\cite{rozanova2023interventional}. Our work is based on mnestic probing to find the features of interest. Here I define task based on the natural logic. Then train task specific probes. Using the trained probes, perform itterative null space projection to remove the surplus features.

\section{Approach}
\begin{figure}[h]
    \centering
    \includegraphics[width=0.70\linewidth]{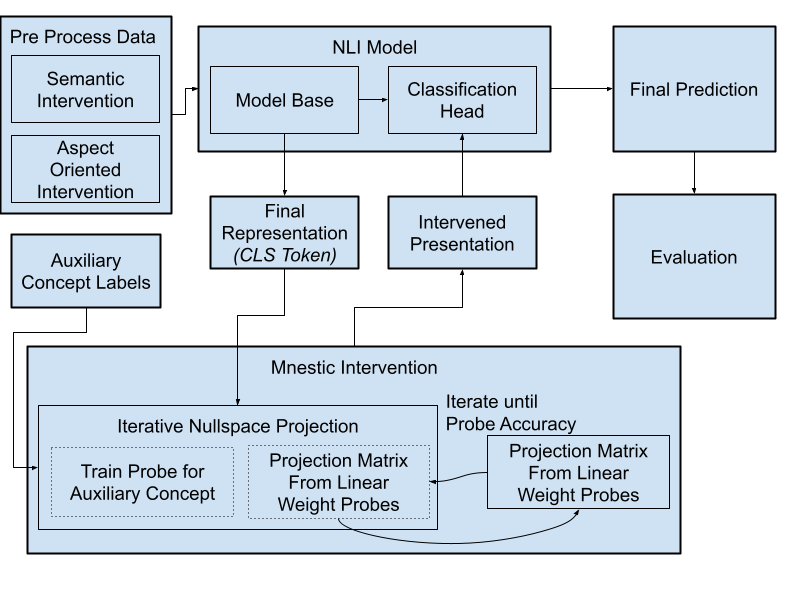}
    \caption{Development approach}
    \label{fig:methodology}
\end{figure}
I use SciFive as base model. The model is used in \textit{A Multi-granularity System for CTR-based Textual Entailment and Evidence Retrieval} as base model developed by Zhou et al. They ranked top most with F-1 score of 85\%. I fine tune the SciFive model on \href{Clinical Trials Data}{https://github.com/ai-systems/Task-2-SemEval-2024}. Define control task for probing. Prepared dataset for sepcific task. Train the task specific probes. Investigate and fine tune the model using probes. Finally, evaluated the fine tunned model. Follwoing sections describes each of the steps in detail. Complete procedure is illustrated in figure \ref{fig:methodology}

\subsection{Data Preprocessing}
\begin{figure}[h]
    \centering
    \includegraphics[width=0.70\linewidth]{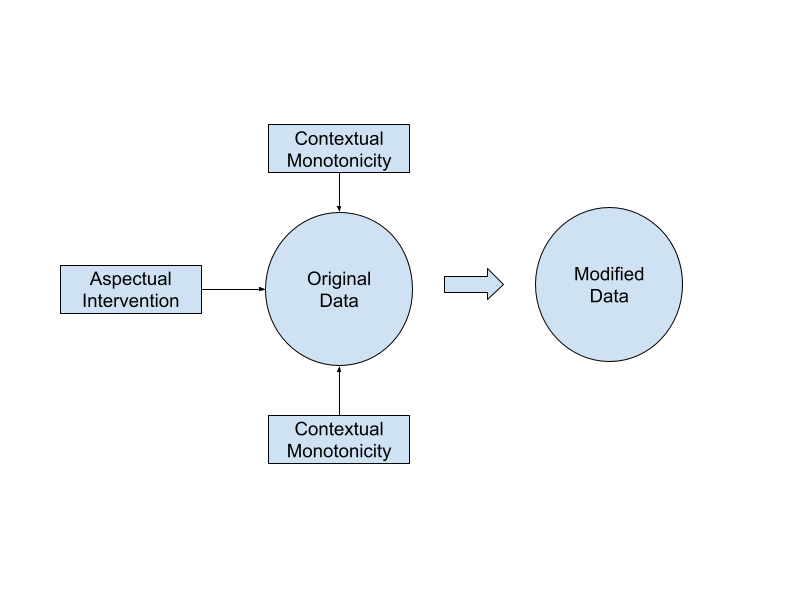}
    \caption{Data Preprocessing}
    \label{fig:preprocessing}
\end{figure}

Context montonicity and Concept relation are two key component of Natural Logic.
\begin{figure}[h]
    \centering
    \includegraphics[width=0.70\linewidth]{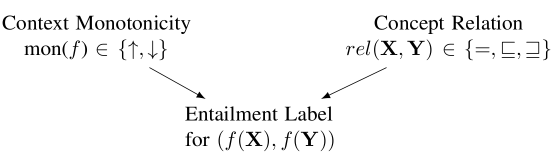}
    \label{fig:natural_logic}
\end{figure}
They form the basis for calculus, determining entailment from substitution. NLI is often described with reference to substitution operation. Affect of substitution may be either forward entailment or reverse entailment or no affect at all.  I intervene the \href{data}{https://github.com/ai-systems/Task-2-SemEval-2024} by considering contextual monotonicity and lexical relationship. Specifically, I rephrase sentences using hyponym, hypernym or unrelated known words. This aspect will help in evaluating semantic correctness of the NLI model. Further, I identify different aspect from data, modify the polarity of aspect, and change the context to investigate the robustness of trained model \ref{fig:preprocessing}.
An example input statement is:
\textit{"Hyperplasia without atypia but with a 10-year modified Gail risk of at least 4\%"}

The given statement is modified for upward monotonicity as follows:

\textit{"Hyperplasia without atypia but with a 10-year modified Gail risk of 9\%"}
Hence, second statement entails the first statement. 

\subsection{Base NLI Model}
\begin{figure}
    \centering
    \includegraphics[width=0.75\linewidth]{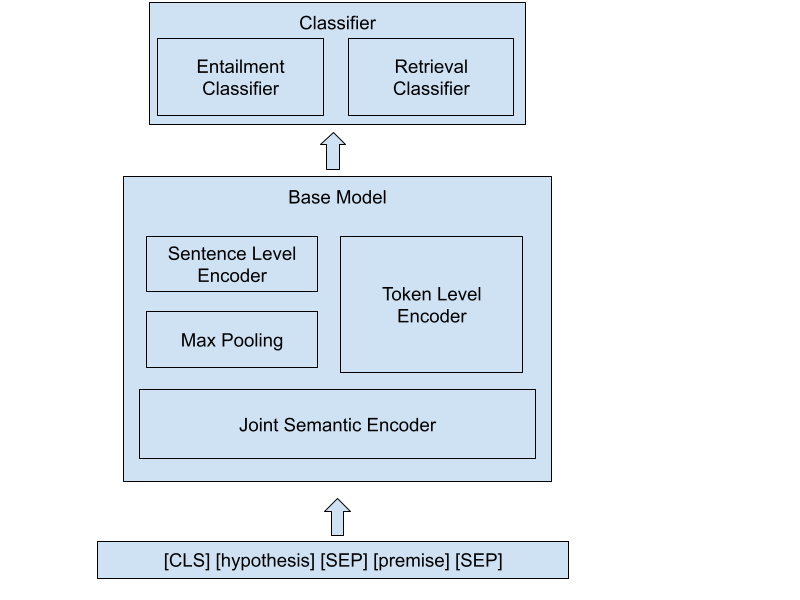}
    \caption{Base Model}
    \label{fig:basemodel}
\end{figure}

I selected the top performer model from SemEval2023 for NLI4CT task. The model \textit{A Multi-granularity System for CTR-based Textual Entailment and Evidence Retrieval} is ranked at top with F-1 score of 85\%\cite{zhou2023thifly}. They developed a combined framework for textual entailment and evidence retrieval tasks \ref{fig:basemodel}. The framework pre process the input statements by adding special tokens. A joint semantic encoder encodes the input sentences. Rest of the network is based on SciFive model. On top of the last decoder layer, final classifier part is added. I remove the final classifier and takes the output of last decoder layer. Pass this output to intervention module. Intervened output from the intervention module is passed again to final classifier for classification.

\subsection{Probing}
Probing is a popular strategy for investigating the presence of intermediate features in neural models. Ferreira et al. proposed a framework to develop prob for semantic interventions \cite{ferreira2021does}. I use the framework to develop and tune the probes. I targeted the context monotonicity task and left concept relation part for future wok. For the task I prepared the custom dataset from the base dataset. Input of the task is a pair of statements and label of the task is entailment and contradiction i.e., whether second statement entails the first statement or contradict the statement. To develop the probe I use simple neural network with one hidden layer. Trained 384 probes of varying complexity on the custom developed dataset. Used these probes in the intervention module for iterative null space projection. Dimension of probes is shown in table\ref{tab:prob_table}.

\subsection{Mnestic Intervention \& Fine Tuning}
The idea is to gradually remove the superfluous features learned during the training and fine tuning of the base model. To accomplish the task, I used the idea of iterative null space projection to update the model parameters while preserving the required features. Null space projection is calculated using trained probes. Nullsapce projection is calculated using equation \ref{eq:nullspace_projection}.
\begin{equation}\label{eq:nullspace_projection}
    \nabla L_{\text{proj}} = \nabla L - P (P^T P)^{-1} P^T \nabla L
\end{equation}
Where \( P \) be the probe's weight matrix, and \( \nabla L \) be the gradient of the loss with respect to the pre-trained model's parameters. \( \nabla L_{\text{proj}} \) is nullspace projection.
I targeted the last layer of decoder part. Reason selecting the last layer is that the earlier layers in the network are known to capture the abstract and granular features whereas as the last layers are know to capture more task specific features. I passed the last layer vectors to intervention module. The intervention module used the trained probes to perform the iterative null space projection. During the process, accuracy is continuously monitored. I performed the iteration for varying number of epochs. The updated vector will be passed again to the model to calculate the accuracy score. If a feature doen’t contribute in the task, it is removed.
\begin{table}
    \centering
    \begin{tabular}{|c|c|c|c|}
     \hline
        \textbf{Probe No.} &\textbf{Input} & \textbf{Hidden} & \textbf{Output}\\ \hline
         1 & 768 & 3 & 1 \\ \hline
         2 & 768 & 6 & 1 \\ \hline
         3 & 768 & 12 & 1 \\ \hline
         4 & 768 & 24 & 1 \\ \hline
         5 & 768 & 48 & 1 \\ \hline
         6 & 768 & 96 & 1 \\ \hline
         7 & 768 & 192 & 1 \\ \hline
         8 & 768 & 384 & 1 \\ \hline
    \end{tabular}
    \caption{Probing layers architecture}
    \label{tab:prob_table}
\end{table}

\section{Evaluation and Experiments}
\subsection{Train base model}
Before fine tuning the model, trained the complete model end-to-end for \(100\) epochs. Training process is performed on Google Colabe T4 TPU. Due to the limited resources only first \(100\) clinical trials are elected for the training process. This resulted in test accuracy of \(91\%\).
\subsection{Training probes}
To train the probes, data is prepared from the base clinical trials data. Handcrafted pair of \(80\) statement. First part contains the original statement from the dataset. Second statement is derived from the first statement. Second statement either entails the first statement or contradict the first statement. Corresponding label is assigned to the statement pair. Each of the probe is trained for \(20\) epochs with a learning rate of \(0.001\). 
\subsection{Fine tune trained model using iterative null projection}
Once all of the probes are trained, I used trained probes to further fine tune the trained model. The process was performed in an iterative manner. Specifically, I retrieved the output of last layer of the decoder (from base model). Used these output as input of the probe. Performed projection on the hidden representation of the trained probes and updated the parameters of the base model. I repeat the process with all of the probes.
\subsection{Evaluate results}
The key evaluation metric for the interventional paradigm is probing accuracy i.e., how before and after null space projection the accuracy is affected. To analyze different aspects of data, I select a particular feature intervening probe one by one. I null out all of the features except the features directed by the selected probe. If it results in degradation of classification accuracy means the feature is required for the classification task and vice versa. Once the model is trained, I evaluate its performance on evaluation dataset. The evaluation results shows that the accuracy of the probe no 9 (with 384) hidden state was 93\%, 2\% higher than the base model.

\section{Conclusion}
LLMs have achieved state-of-the-art performance in many NLP fields like Natural Language Generation \& Understanding, Language Translation, Information retrieval. But there limitation regarding  factual inconsistency and hallucination has put question on their usage. There is adverse degradation in performance if there is a little data shift. The use of LLMs in vital fields like real-world clinical trials necessitates the creation of innovative evaluation methods rooted in more methodical behavioral and causal analyses \cite{xing2020tasty}. In this work I've proposed a probing based technique to fine tune the model and enhance its integrity. Results shows that resulting model is more robust and perform decision with definite reason.
\section{Limitations \& Future Work}
Due to the limitation of available compute resources, I've elected \(100\) trials from total of \(1710\) trials in the given dataset. I've trained only 8 probes for monotoncity tasks. The number is much lower as suggested by the developer of Probe-ably framework. In their work they trained 50 probes for each of the task. Also, I've considered only monotonicity part and doesn't consider Concept relation part. They need to be investigated in the future, as conceptual relation is the key component of natural logic.

\end{document}